\documentclass[lettersize,journal]{IEEEtran}
\usepackage{amsmath,amsfonts}
\usepackage{algorithmic}
\usepackage{algorithm}
\usepackage{array}
\usepackage[caption=false,font=normalsize,labelfont=sf,textfont=sf]{subfig}
\usepackage{textcomp}
\usepackage{stfloats}
\usepackage{url}
\usepackage{verbatim}
\usepackage{graphicx}
\usepackage{cite}
\usepackage{multirow}
\usepackage{xcolor}
\begin{document}

\title{On-device Federated Learning in Smartphones for Detecting Depression from Reddit Posts}

\author{Mustofa Ahmed, Abdul Muntakim, Nawrin Tabassum, Mohammad Asifur Rahim, and \\ Faisal Muhammad Shah
\thanks{Mustofa Ahmed, Nawrin Tabassum, and Faisal Muhammad Shah are with the Department of Computer Science and Engineering, Ahsanullah University of Science and Technology, Dhaka 1208, Bangladesh (e-mail: mustofa.cse@aust.edu; nawrin.cse@aust.edu; faisal.cse@aust.edu).}
\thanks{Abdul Muntakim is with the Department of Computer Science and Engineering, Khulna University of Engineering \& Technology, Khulna-9203, Bangladesh (e-mail: basitmuntakim@gmail.com).}
\thanks{Mohammad Asifur Rahim is with the
Department of Information and Computer Science,
King Fahd University of Petroleum and Minerals(KFUPM), Dhahran, Saudi Arabia (e-mail: g202417400@kfupm.edu.sa).
}
}





\maketitle

\begin{abstract}
 Depression detection using deep learning models has been widely explored in previous studies, especially due to the large amounts of data available from social media posts. These posts provide valuable information about individuals' mental health conditions and can be leveraged to train models and identify patterns in the data. However, distributed learning approaches have not been extensively explored in this domain. In this study, we adopt Federated Learning (FL) to facilitate decentralized training on smartphones while protecting user data privacy. We train three neural network architectures--GRU, RNN, and LSTM on Reddit posts to detect signs of depression and evaluate their performance under heterogeneous FL settings. To optimize the training process, we leverage a common tokenizer across all client devices, which reduces the computational load. Additionally, we analyze resource consumption and communication costs on smartphones to assess their impact in a real-world FL environment. Our experimental results demonstrate that the federated models achieve comparable performance to the centralized models. This study highlights the potential of FL for decentralized mental health prediction by providing a secure and efficient model training process on edge devices.
\end{abstract}

\begin{IEEEkeywords}
Federated learning, text classification, depression detection, social media, smartphones.
\end{IEEEkeywords}

\section{Introduction}

Deep Learning has been remarkably successful in recent years owing to its ability to learn and extract patterns from large amounts of data, whereas traditional machine learning algorithms often can not detect useful patterns from these data. Deep Learning has been particularly effective in Natural Language Processing (NLP) tasks, for instance, sentiment analysis \cite{tan2022roberta,zhang2023empirical}, automatic summarization \cite{ma2021t}, and text classification \cite{manias2023multilingual}. Recently, it has seen widespread use in the healthcare domain for various applications, ranging from medical diagnosis \cite{yan2023towards} to clinical data analysis \cite{tang2022self}. Social media sites-- Facebook, Twitter, and Reddit generate large amounts of text data containing information about individuals' feelings and emotions \cite{di2023methodologies}. Although data from these sources are often unstructured, deep learning algorithms can be leveraged to examine them and identify any early signs of mental illness \cite{uban2021emotion,ilias2023calibration}.

Depression is a critical mental health condition characterized by continuous feelings of sadness, hopelessness, and loss of interest in regular activities \cite{chikersal2021detecting}. It can significantly affect an individual’s everyday life and degrade their overall quality of life. Early diagnosis of depression can not only facilitate timely intervention but also prevent adverse outcomes. Deep learning models can play a pivotal role in identifying signs of depression from user data.
Social media texts often reflect a person's emotional state and offer insights into their daily life activities. A person with signs of depression might use negative language while expressing feelings of personal struggles, loneliness, and disappointment \cite{bathina2021individuals}. These features can be challenging to recognize for conventional machine learning algorithms. However, deep learning algorithms are specifically designed to analyze such complex and unstructured data and are highly effective when dealing with data collected from social media posts \cite{kodati2023identifying}.  

There are significant concerns associated with the privacy of social media data as they contain sensitive and personal information, including users’ thoughts and behaviors. Traditional machine learning requires data to be collected in a central server, which establishes risks of data breaches, misuse of information, and unauthorized access. Besides, users might feel uncomfortable with their data being stored and analyzed. These challenges highlight the need for privacy-preserving approaches to machine learning and deep learning that address the ethical concerns related to the use of such data. Assuring the security and privacy of social media data is particularly important as the disclosure of these data might result in unintended harm and consequences. Regulatory standards, namely the General Data Protection Regulation (GDPR) \cite{voigt2017eu}, should be strictly followed by the organizations that develop machine learning models to assure user data privacy throughout the training process. 
\IEEEpubidadjcol

Federated Learning (FL) offers a promising solution to the data privacy concerns surrounding the traditional model training process \cite{li2020federated}. FL allows collaborative training on decentralized data and eliminates the need to transfer data to a central server. User data remains on edge devices, such as smartphones, tablets, and IoT devices. Each client trains the local model on its own dataset and shares the model updates or gradients with a central server, which aggregates the gradients to update the global model \cite{mcmahan2017communication,pillutla2022robust}. Raw data are not communicated to the server, reducing the inherent privacy risks related to data centralization. FL performs decentralized training on valuable data residing in distributed data sources to improve model performance and protect data privacy simultaneously \cite{choudhury2019differential}. This balance is crucial to enable models to learn from real-world data that are often distributed and privacy-sensitive.

The application of Federated Learning has shown significant potential, especially in mental health diagnostics \cite{xu2021privacy,tabassum2023depression}, where data privacy must be ensured. Although deep learning can be an effective solution, the highly sensitive nature of personal user data requires a robust privacy-preserving approach. However, protecting the privacy of social media data can be challenging when building a depression detection model. Federated Learning addresses the challenges by ensuring that users' private data remain confidential while enhancing the global model’s ability to accurately detect depression.
In addition, FL addresses the issue of data heterogeneity, which is prevalent in real-world healthcare data \cite{liu2022contribution,nguyen2022federated}. It enables model training on diverse datasets from varied sources, which improves the model’s generalizability and robustness \cite{li2020fedopt}. 

In this paper, we present a federated learning based model training approach for detecting signs of depression from Reddit posts. We develop a system with smartphones as local devices where private data are securely stored. Additionally, we evaluate the training time, overhead time, upload-download time, and inference time of the models on mobile devices. Previous works mainly focused on simulating the FL process using smartphone emulators \cite{esteves2023towards,jiang2022flsys}\ and mostly performed image classification tasks. In contrast, our work focuses on an NLP task and trains local models on text data stored on smartphones. We use three neural network architectures for on-device training and compare their performance and applicability for privacy-preserving depression detection using federated learning. Along with conducting a comprehensive performance analysis of our proposed method in various FL scenarios, we highlight the communication costs between edge devices and the server. This study provides insights into how FL can be implemented effectively in real-world environments while evaluating model accuracy, resource consumption, and communication efficiency.

This study provides the following key contributions:
\begin{itemize}
    \item We propose a federated learning method for training a depression detection model using Reddit post texts stored on mobile devices
    \item We employ three neural network architectures--GRU, RNN, and LSTM for training and demonstrate comparable performance under various data distributions
    \item We propose a novel approach that uses a common tokenizer across all client devices for enhanced model performance and reduced computational load
    \item We analyze resource consumption and communication costs of various models on smartphones to assess their impact on the real-world federated learning process
\end{itemize}

\section{Related Works}

Federated Learning architectures have been widely studied across various NLP domains \cite{zhang2023fedlegal} because of the increasing concerns over users' data privacy. Khalil et al. \cite{khalil2024federated} present a novel application of FL for privacy-preserving depression detection using multilingual social media data. The methodology involved four experimental setups: local training on individual monolingual datasets, centralized training combining all datasets, FL with IID data, and FL with non-IID data, where each client holds data in a different language. Li et al. \cite{li2023intelligent} introduced the CNN-based CAFed algorithm for detecting depression using FL. CAFed addresses non-convex optimization problems, achieving faster convergence and reduced communication overhead compared to FedAvg, especially in resource-constrained and heterogeneous environments.

The majority of federated learning works are usually based on simulation \cite{liu2021learning}. However, several studies have implemented federated learning for smartphone devices. One of the first implementations was done by Google \cite{hard2018federated}, where the authors tried to predict the next keyword on Android devices. However, no explicit details regarding implementation were mentioned apart from stating that Tensorflow Lite \cite{googleLiteRTOverview} was employed to create the app. Bn and Abdullah \cite{bn2022privacy} have predicted whether a person is depressed through speech analysis. A model trained on a centralized setup was deployed to the mobile devices through Tensorflow Lite. Additionally, they showed details of memory usage, inference time, and energy consumption of Android devices with the help of Android profiling. Esteves et al. \cite{esteves2023towards} have performed image classification on Cifar-10 and Cifar-100 using Android emulators. Nevertheless, no details concerning Android implementation were mentioned. Suruliraj and Orji \cite{suruliraj2022federated} developed a federated learning app for iOS devices using the CoreML library that monitors sensor data from users' phones and tries to predict if a person is depressed by identifying unusual patterns in daily activities. Despite showing memory, RAM, and power usage for running the app, no deep learning models were trained. Instead, only an anomaly detection algorithm was used. Michalek et al. \cite{michalek2023proposal} suggested that a federated learning app could be built using Tensorflow Lite for clients, and servers could be built for reliable communication using Apache Kafka and Firebase Realtime Database. Jiang et al. \cite{jiang2022flsys} have developed a federated learning Android app using  Deep Learning for Java (DL4J), through which they conducted extensive simulations for Human Activity Recognition and Sentiment Analysis using the app. Tabassum et al. \cite{tabassum2023depression} also built an app using DL4J to predict depression severity in a person using smartphone sensor data. 

Several frameworks, such as Flower \cite{mathur2021device}, Hermes\cite{li2021hermes}, FlaaS\cite{katevas2022flaas}, and FedScale\cite{lai2022fedscale}, support federated learning on Android devices. Each of these frameworks has unique support, such as enhanced communication techniques, support for asynchronous systems, advanced federated learning strategies, and many more. These frameworks must be paired with TensorFlow Lite or Pytorch Mobile \cite{pytorchHome} for on-device training. TensorFlow Lite and Pytorch Mobile are optimized for performance on constrained mobile devices. However, they introduce some abstraction and restriction, making complete control over models, architecture, and strategies challenging. 

We chose the Chaquopy \cite{chaquoEasiestPython} Plugin to build our federated learning app, as it allows us to run Python code natively on Android devices. Chaquopy simplifies the workflow, providing complete control over customized model training without the constraints imposed by mobile-specific libraries. Essentially, Chaquopy offers greater flexibility in implementing custom federated learning architectures, customized communication protocols, and aggregation strategies without being tied to the predefined structure or protocols of frameworks.

\section{Preliminaries}

\subsection{Deep Learning}

In our study, we use three neural network models: Recurrent Neural Network (RNN), Gated Recurrent Unit (GRU), and Long Short-Term Memory (LSTM). These models have been selected because they can capture temporal dependencies, which makes them effective in understanding the contextual nature of language when predicting depression severity.

An RNN is a Neural Network where neurons in a layer are interconnected with neurons of previous layers or the same layer in a loop. RNNs maintain a hidden state as they process each element in a sequence, which helps them capture temporal dependencies. However, standard RNNs suffer from the vanishing gradient problem that limits their ability to learn long-term dependencies. The GRU is an evolution of the traditional RNN that addresses the vanishing gradient issue. It uses two gating mechanisms, the update gate and the reset gate, that control the information flow and assist it in capturing longer dependencies compared to RNN. LSTMs improve upon RNNs by including a more sophisticated gating mechanism. An LSTM cell has an input gate, a forget gate, and an output gate. These three gates collectively allow LSTMs to manage long-term dependencies effectively, making LSTM robust for complex sequential tasks like depression prediction.

\subsection{Federated Learning}

Federated Learning (FL) \cite{li2020fedopt,mcmahan2017communication} is a distributed machine learning method that allows decentralized training on local devices. The private training data are located on edge devices and never sent to the central server. A major advantage of this approach is enhanced data privacy and security, as it does not require data centralization. Clients train a global model collaboratively by only transferring their local model updates to a central server. The local weights and biases are aggregated on the server and transmitted to the clients for the next round of training. In this way, FL operates in multiple rounds. FL is suitable for utilizing the computational resources of edge devices for model training. Additionally, FL ensures communication efficiency by optimizing the size of shared weights and biases. However, FL faces unique challenges in real-world deployment, such as data and system heterogeneity and privacy attacks \cite{yang2024fedfed,ilhan2023scalefl,tabassum2024efficiency}.   

\subsubsection{FedAvg}

FedAvg (Federated Averaging) \cite{mcmahan2017communication} is the most widely used technique in Federated Learning that works by combining the local model updates received from the distributed data sources. The main idea is to perform a weighted averaging of the weights and biases from a subset of clients and update the global model for the next round of training. The weighted average ensures that the clients with larger datasets have a larger impact on the global model. Let $N$ be the total number of clients and $C$ = \{$C_{1}, C_{2}, C_{3}, ….., C_{N}$\} be the set of clients participating in the FL process; each client holding a local dataset \{$D_{1}, D_{2}, D_{3}, ……, D_{N}\}$. In round $t$, a subset of $K$ clients $c \in C$ receives the global model parameters $W_{t}$ and trains the local model $W_t^c$ by minimizing the loss function $L_c(W)$. Finally, the server receives the local models and performs FedAvg as 
\begin{equation}
    W_{t+1} = \sum_{k=1}^{K} \frac{\left| D_k \right|}{\left| D \right|} W_t^k
\end{equation}
where $\left| D_k \right|$
denotes the data size of client $k$ and $\left| D \right|$ denotes the total data size across all clients. The goal of the FedAvg algorithm is to minimize the following objective over multiple rounds until the model converges.
\begin{equation}
       \underset{W}{min \:}   \sum_{k=1}^{N} \frac{\left| D_k \right|}{\left| D \right|} L_{k}(W)
\end{equation}

\subsubsection{IID Vs Non-IID}

IID (Independent and Identically Distributed) and non-IID (Non-Independent and Identically Distributed) data distributions are two common phenomena in Federated Learning that impact the performance of both global and local models. The data across all participants come from the same underlying distribution in an IID setting, whereas the data come from different distributions in a non-IID setting. Generally, two types of imbalances occur in a non-IID setting-- data imbalance and class imbalance.

In case of data imbalance, the amount of data varies across different clients. Let $|D_i|$ be the number of samples that belong to the local dataset of client $i$. The data imbalance can be formulated as 
\begin{equation}
\left| D_i \right| \neq \left| D_j \right| , \quad \exists i, j : i \neq j \end{equation}
In case of class imbalance, the proportion of class labels varies across different clients. Some clients may hold datasets where most samples belong to a certain class, and the remaining classes may have little or no representation in the dataset. Let $Q$ be the set of all class labels and $P_{i}^q$ be the proportion of samples from a class $q$ that belong to the dataset of client $i$. The class imbalance can be formulated as 
\begin{equation}
    P_i^q \neq P_j^q ,  \quad \exists i, j : i \neq j \text{ and } q \in Q
\end{equation}
Class imbalance results in poor generalization and degraded model performance. Local models trained on datasets with imbalanced class distribution may be biased and overfitted to the majority class. Moreover, local models may not classify the underrepresented classes correctly, resulting in reduced accuracy of the global model. In this study, we perform experiments in both IID and non-IID (class imbalance) settings to demonstrate the effectiveness of our method in a real-world data heterogeneous setup.

\section{Dataset}

\subsection{Dataset Description}
Numerous researchers have conducted studies on detecting depression from social media texts which primarily focus on whether a person is depressed or not \cite{wolohan2018detecting, tsugawa2015recognizing}. For our research, we utilized Sampath and Durairaj's dataset \cite{sampath2022data}, as it aims to assess a person's depression severity instead of just diagnosing depression. The dataset comprises English posts from Reddit and includes three labels: Severely depressed, Moderately depressed, and Not depressed.

The dataset includes a train set, a dev (development) set, and a test set. The train set consists of 8891 samples, 2720 of which are unique. The dev set consists of 4496 samples, with 4481 being distinct. As there are many duplicate instances, we used the preprocessed variant of the dataset created by Poświata and Perełkiewicz \cite{poswiata2022opi}. In this preprocessed dataset, the training and dev sets were merged after removing duplicates, resulting in a final dataset of 7006 samples, with 6006 allocated to the training set and 1000 reserved for the dev set. Lastly, the number of test instances in the dataset was 3245.

\subsection{Data Augmentation}

\begin{figure}[!t]
\centering
\includegraphics[scale=0.55]{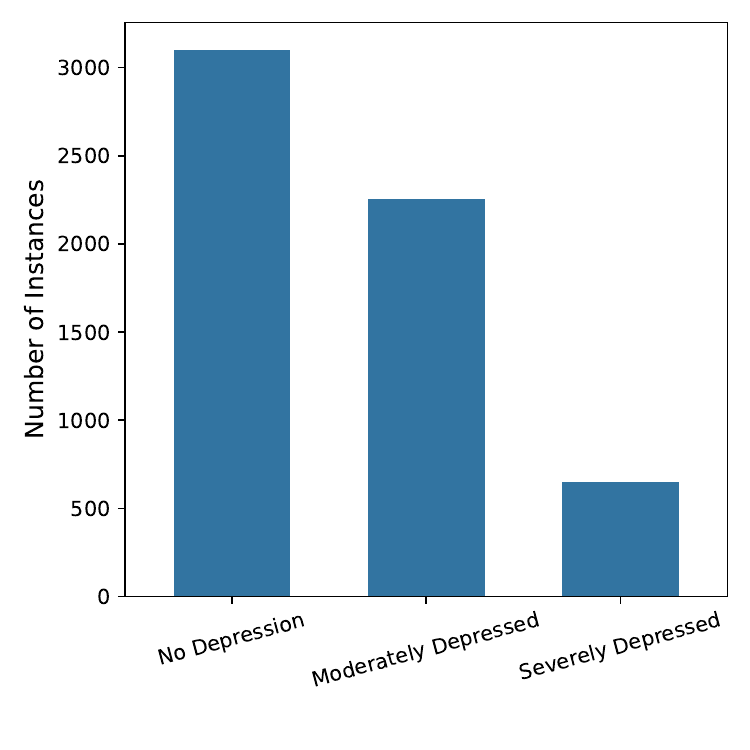}
\caption{Class Distribution of the Imbalanced Dataset}
\label{fig1}
\end{figure}

Figure \ref{fig1} highlights the imbalance in the training dataset, which can lead to various performance issues. To overcome this problem, we utilized a word-level substitution strategy using word embedding to augment the data. This augmentation strategy selects random words from a sample text and substitutes them with a semantically similar word. We leveraged the GloVe 840B.300d model for this task as it has an extensive vocabulary and can accurately locate similar words. By applying this augmentation, we increased the minority class instances to create a balanced dataset with total instances standing at 9303.

\subsection{Data Shards}

Initially, we intended to collect text data from the five client devices. However, training a deep learning model demands a large dataset, and ensuring the validity of the data requires considerable work. Thus, we selected an existing, well-established dataset that offers credibility and enables us to compare and assess our model’s performance with other research studies. The dataset was evenly divided among the five clients. For the IID distribution of the dataset, all clients received a total of 1860 instances, and from each class, 620 samples. Table \ref{tab:table1} shows the non-IID distribution of how each of the three classes is distributed amongst the client devices.

\begin{table}[]
\centering
\caption{Non-IID Class Distribution Across Clients}
\label{tab:table1}
\begin{tabular}{|c|c|c|c|}
\hline
Clients  & \begin{tabular}[c]{@{}c@{}}Severely \\ Depressed (\%)\end{tabular} & \begin{tabular}[c]{@{}c@{}}Moderately \\ Depressed (\%)\end{tabular} & \begin{tabular}[c]{@{}c@{}}Not \\ Depressed (\%)\end{tabular} \\ \hline
Client 1 & 40                                                            & 10                                                              & 10                                                       \\ \hline
Client 2 & 10                                                            & 10                                                              & 40                                                       \\ \hline
Client 3 & 10                                                            & 40                                                              & 10                                                       \\ \hline
Client 4 & 10                                                            & 30                                                              & 20                                                       \\ \hline
Client 5 & 30                                                            & 10                                                              & 20                                                       \\ \hline
\end{tabular}
\end{table}

\section{Methodology}

\subsection{Data Processing}
\subsubsection{Padding}

Padding equalizes the input sequence lengths for our model. In our case, we set the maximum length of the input sequence to be 100 tokens, which also includes extending the sequences that do not reach this maximum length by appending zeros. The padding ensures all the input vectors presented to the model have the same uniform size. This step is essential because the architecture of our neural networks requires the input data to have the same shape. We experimented with maximum lengths of 150 and 200 as well but found no noticeable improvement in performance.

\subsubsection{Embedding}

In word embeddings, individual words are represented as real-valued vectors in a vector space. This technique allows for some measure of similarity between words, effective computation, and utilization of semantic relationships. Embedding has revolutionized how machines understand human language.

GloVe (Global Vectors for Word Representation) is an unsupervised learning algorithm for generating word embeddings by aggregating global word-word co-occurrence statistics from a corpus. The model is designed to leverage the overall statistics of word occurrences in a dataset to produce a space where the distance between any two words relates to their semantic similarity. The objective function of GloVe in Formula \ref{glove_eqn} is designed to minimize the difference between the logarithm of their co-occurrence count and the dot product of word vectors. 

\begin{equation}
    G = \sum_{i,j}f(X_{ij})(w^T_i\cdot\Tilde{w}_j+b_i+\Tilde{b}_j-log(X_{ij}))^2
    \label{glove_eqn}
\end{equation}

 where $w_i$ and $b_i$ are the word vector and bias respectively of word $i$, $\tilde w_j$ and  $\tilde b_j$ are the context word vector and bias respectively of word $j$, $X_{ij}$ is the number of times word $i$ occurs in the context of word $j$, $f$ is a weighting function that assigns lower weights to rare and frequent co-occurrences.
 
GloVe combines the advantages of two major approaches to word embedding: matrix factorization and local context window methods. GloVe embeddings enable algorithms to calculate semantic similarity between words or phrases and address analogy-based tasks, which makes this method suitable for diverse datasets and when capturing global context is important.

In our experimentation with GloVe, we tested different vector dimensions, including 50-dimensional (50d), 100-dimensional (100d), and 200-dimensional (200d) vectors. We found that the 100d and 200d vectors performed significantly better than the 50d vectors. The increase in dimensions allows a more detailed semantic capture as it can encapsulate more information about each word. However, the performance between 100d and 200d was nearly identical, so we used 100d due to lower computational resource requirements.

\subsection{Common Tokenizer}

Tokenization is one of the rudimental steps of text preprocessing. It helps break down complex sentences into manageable units and standardized tokens. It also makes it easier to handle diverse input lengths and structures, which can be further analyzed to identify patterns in the text.

In the centralized deep learning approach, a tokenizer can be trained on the whole dataset. The tokenizer processes and generates a consistent vocabulary from the local corpus, and all instances of text data are processed using the same tokenizer. This existing strategy is not applicable in FL because the data is distributed. One possible strategy is for clients to train their own tokenizers. In this case, differences in vocabulary can arise due to diverse local datasets in the FL setup. This approach causes mismatches and problems with unknown words during model aggregation, which leads to poor performance of the deep learning models.

In our work, we propose a common tokenizer for all clients in a FL configuration to address this issue. Instead of allowing clients to train their individual tokenizer, we provide a common tokenizer to all the clients. The common tokenizer was created by training it on an existing corpus by Poświata and Perełkiewicz \cite{poswiata2022opi}. The corpus contains Reddit Mental Health Dataset posts and two mental health-related subreddits \cite{low2020natural}. This step ensures the clients and server use the same vocabulary and tokenization rules, which dismisses problems caused by clients training their individual tokenizers and assists in processing text data uniformly across all clients.

\begin{figure}[!t]
\centering
\includegraphics[width=\linewidth]{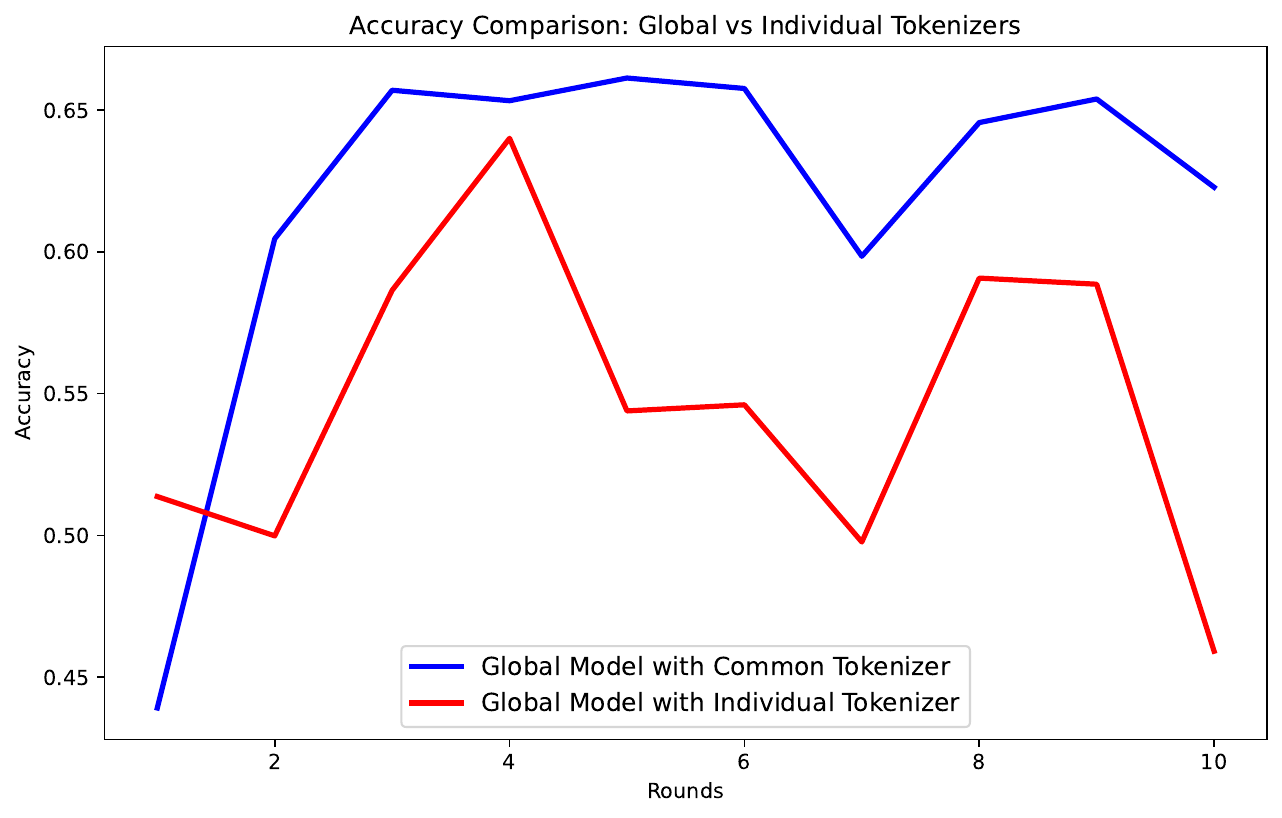}
\caption{Comparison of RNN Model Accuracy Using a Common Tokenizer vs. Client-Trained Tokenizers in a Federated Learning Setup}
\label{fig_tokenizer}
\end{figure}

The result in Figure \ref{fig_tokenizer} shows the significance of a common tokenizer in federated settings. A common tokenization strategy provides faster convergence, higher accuracy, and reduced performance variance across clients compared to client-specific tokenizers. The common tokenizer approach also ensures quicker achievement of optimal performance by ensuring lower overhead, which we found experimentally.

\subsection{Firebase}
Robust, scalable storage and real-time data management are essential for handling complex processes like a FL system. Firebase is a suitable platform for FL as it facilitates seamless communication and offers efficient and reliable data storage to maintain model changes. Furthermore, it provides support for synchronizing data across several client devices.

\subsubsection{Firebase Storage}
We used Firebase Storage (FS) to save model parameters, which include weights and biases, from the clients participating in the federated learning process. FS is a reliable platform for storing these model parameters. After each training session, the client devices upload their model parameters to FS, where the server can access the parameters seamlessly and generate the updated global model. Similarly, the client devices can retrieve the updated global parameters effectively.

\subsubsection{Firebase RealTime Database}
The Firebase Realtime Database (FRD) is crucial in handling the real-time synchronization messages between the client devices and the server in our FL setup. It assists in tracking the status of both clients and the server by storing critical information, such as when clients complete local training, upload their local model parameters, and when the server uploads the updated global model parameters. 

\subsection{Model Architecture}

In our study, we employed RNN, GRU, and LSTM models. The initial layer of all the models is a non-trainable embedding layer that utilizes the pre-trained GloVe embedding to map the input token to a 100-dimensional vector. In the first model, an RNN layer of 400 units with a ReLU activation function follows the embedding layer to capture the sequential dependencies of the text data. Next, a Dense layer of 300 units with a Relu activation function condenses the features extracted by the RNN layer. To mitigate overfitting, we added a dropout layer with a drop rate of 0.25 after both the RNN and Dense layers. The final layer is a dense layer comprising three units corresponding to the number of target classes. A sigmoid activation function is applied, making the architecture suitable for multi-label classification. We tried various combinations of units for the RNN and Dense layer and found that 400 and 300 units resulted in a model that converged in fewer rounds. In the second and third models, the RNN layer was replaced by GRU and LSTM layers, respectively, while the remaining architecture was unchanged. We trained each local model for one epoch to avoid overfitting and reduced computational load. We opted to use the Adam optimizer due to its efficiency, which makes it ideal for training complex models with faster convergence. Lastly, the learning rate and batch size were set to 0.001 and 32, respectively. 

\subsection{Work Approach}

Algorithm \ref{algo1} shows the client-server communication and the global and local model training process of our proposed approach. Initially, we distributed the data among five clients, providing each with both IID and non-IID data options. The clients can choose the data distribution and shard to train their local model during training. As specified earlier, all clients use a common tokenizer to convert their text to vectors. Afterwards, the data length is also adjusted to make the data equal in length through sequence padding. The clients now have access to the processed data and embedding matrix, ready to train a deep-learning model. Prior to training, the clients must load the weights and biases of the global model, initialized by the server. Once training on the client devices is complete, the local model parameters are uploaded to FS, along with necessary synchronization messages to the FRD. The client device then periodically checks if the server has produced an updated global model.

\begin{figure*}[!t]
\centering
\includegraphics[scale=0.5]{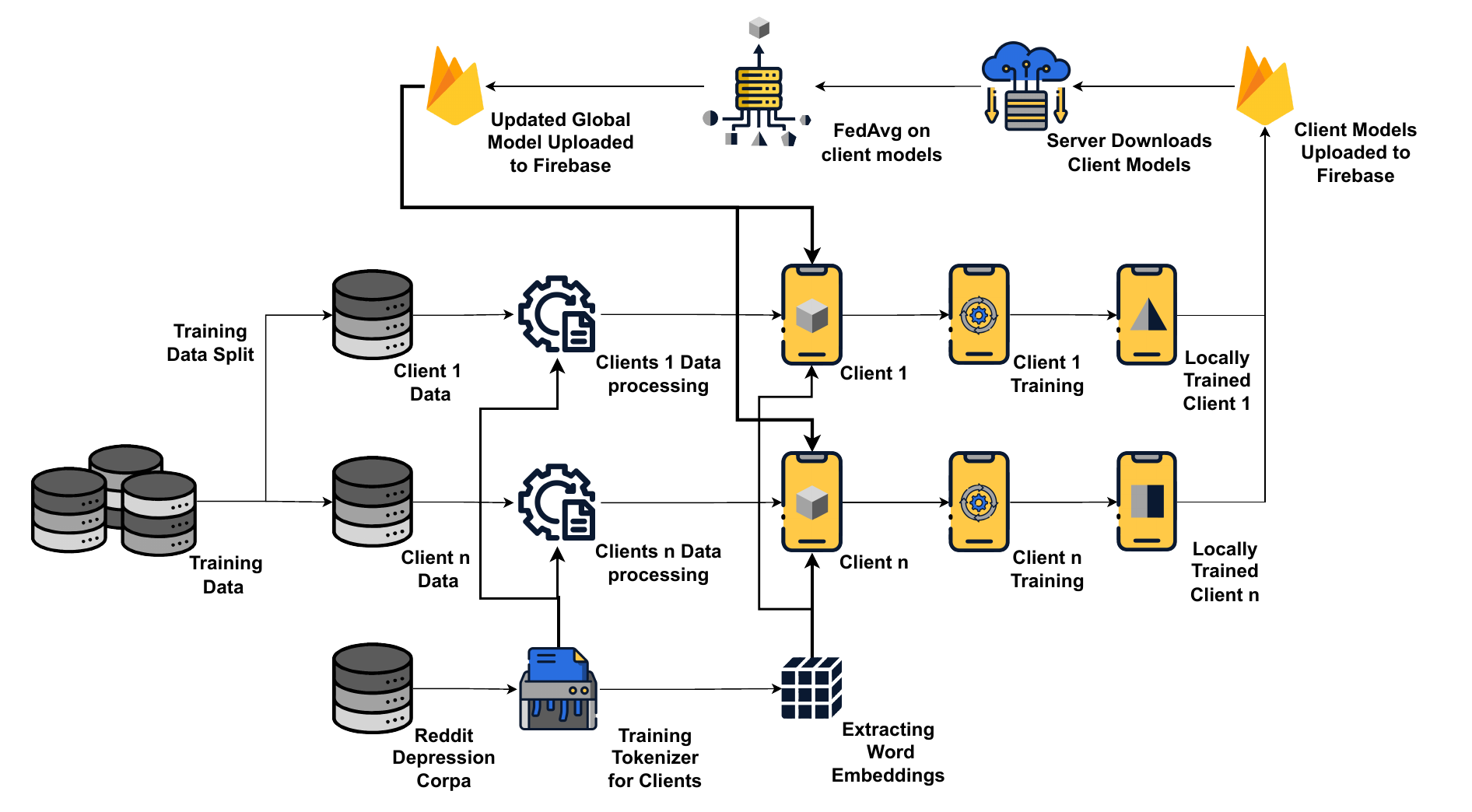}
\caption{Workflow of the Proposed Federated Learning System}
\label{fig3}
\end{figure*}

The server routinely monitors the FRD to track the progress of the local models. Once all client devices have completed their training and uploaded their parameters, the server downloads the client's model parameters. The client models are aggregated using Federated Averaging to produce an updated global model. No testing was performed on the client devices to avoid the additional burden of handling 3245 data instances. Instead, both the global and all the local models are evaluated on the server while the server waits for the client devices to finish training. The server uploads the global model to FS and essential synchronization messages to FRD. The client retrieves the new global model parameters, and the entire process is repeated for ten rounds. A visual representation of our work approach is shown in Figure \ref{fig3}. 

\begin{algorithm}
\caption{Client-Server Communication \& Training}
\label{algo1}
\textbf{Input}: Total rounds $T$, total clients $N$, clients per round $K$, global model $W^0$, local datasets $D_1, D_2, D_3,...,D_N$, learning rate $\eta$.  \\
\textbf{Output}: Trained global model $W_g$. \\
\begin{algorithmic}[1]
\STATE Initialize $W^0$ with random weights
\FOR{$t = 1$ to $T$}
\STATE // \textbf{Client}
\FOR{$k = 1$ to $K$}
\STATE Process $D_k$ 
\STATE Download global model $W_g^{t-1}$ from Firebase
\STATE Train local model $W_k^{t}$ 
\STATE $W_k^{t} = W_g^{t-1} - \eta \nabla W_k^{t-1}$
\STATE Upload $W_k^{t}$ to Firebase 
\STATE Upload synchronization message $M_k^t$ to Firebase 
\ENDFOR
\STATE // \textbf{Server}
\FOR{$k = 1$ to $K$}
\STATE Download local model $W_k^{t}$ from Firebase
\ENDFOR
\STATE Aggregate local models and update global model $W_g^t$
\STATE $W_g^t = \sum_{k=1}^{K} \frac{\left| D_k \right|}{\left| D \right|} W_k^{t}$
\STATE Upload ${W}_g^t$ to Firebase 
\STATE Upload synchronization message $M_g^t$ to Firebase
\ENDFOR
\STATE \textbf{return} $W_g$
\end{algorithmic}
\end{algorithm}

\section{Experimental Results}

\subsection{Setup}

\subsubsection{Client}
The Android app was built using Android Studio (2022.1.1) and requires a minimum Android version of 5.1 (Lollipop) or higher. By leveraging the Chaquopy (14.0.2) plugin, we integrated Python code into the Android app. Chaquopy enables the inclusion of Python libraries from the Python Package Index (PyPI), allowing us to incorporate Python libraries, such as Tensorflow (2.1.0), Pandas (1.3.2), NumPy (1.19.5), and Pyrebase (4.7.1). These libraries were essential in building our federated learning app.

\begin{figure}[!t]
\centering
\includegraphics[scale=0.25]{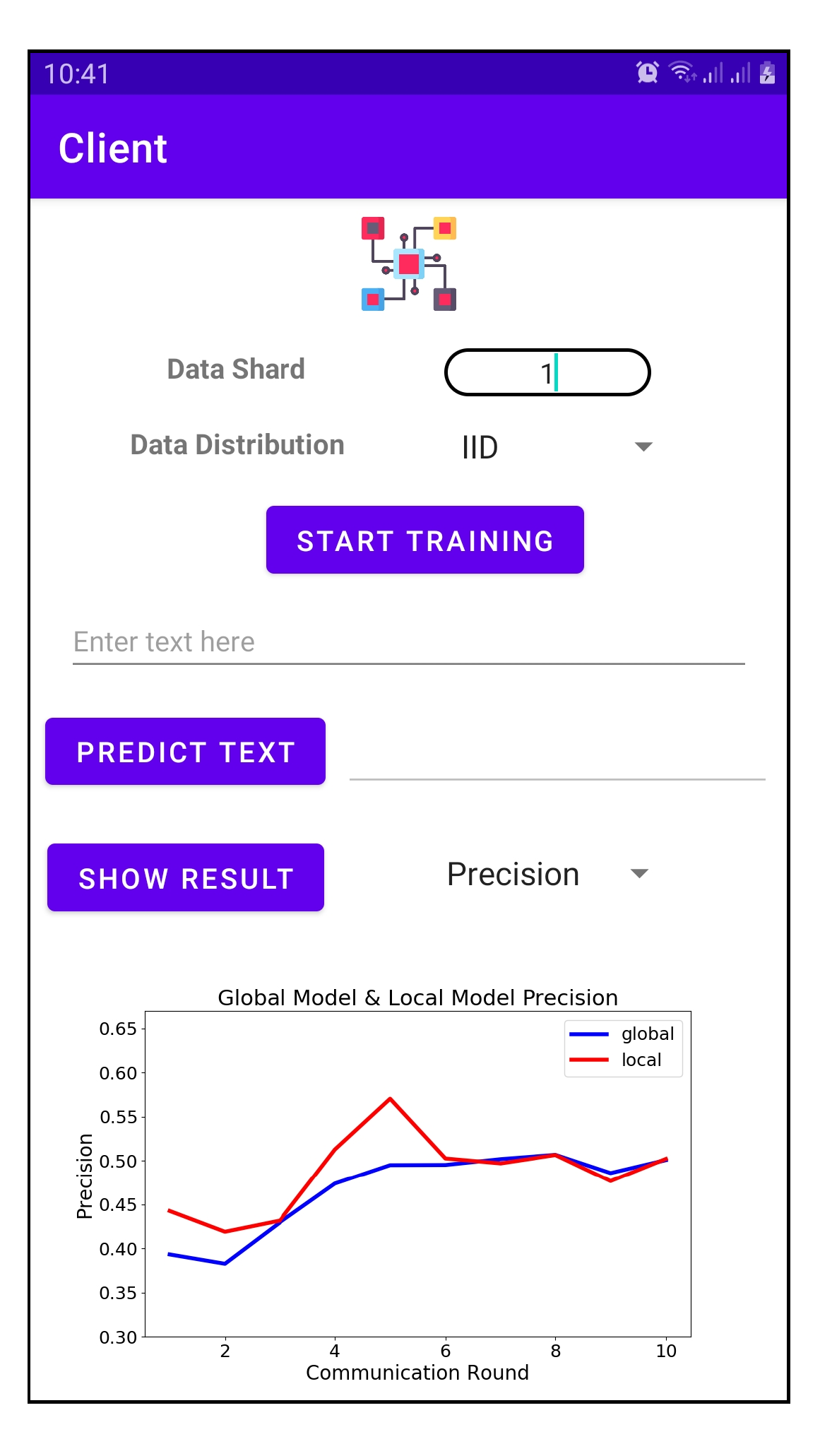}
\caption{User Interface of the Federated Learning Android App}
\label{fig4}
\end{figure}

Clients' devices can select the data shard and distribution to train the deep learning model. The data shard number determines the client's number. To view the training results, clients can choose the evaluation metric, and the outcomes for both the global model and the current client's model will be displayed in a graphical figure at the bottom of the screen. Additionally, the user can input a message, and the app will predict its corresponding label. A screenshot of our Android app is shown in Figure \ref{fig4}.

\subsubsection{Server}
\begin{figure}[!t]
\centering
\includegraphics[width=\linewidth]{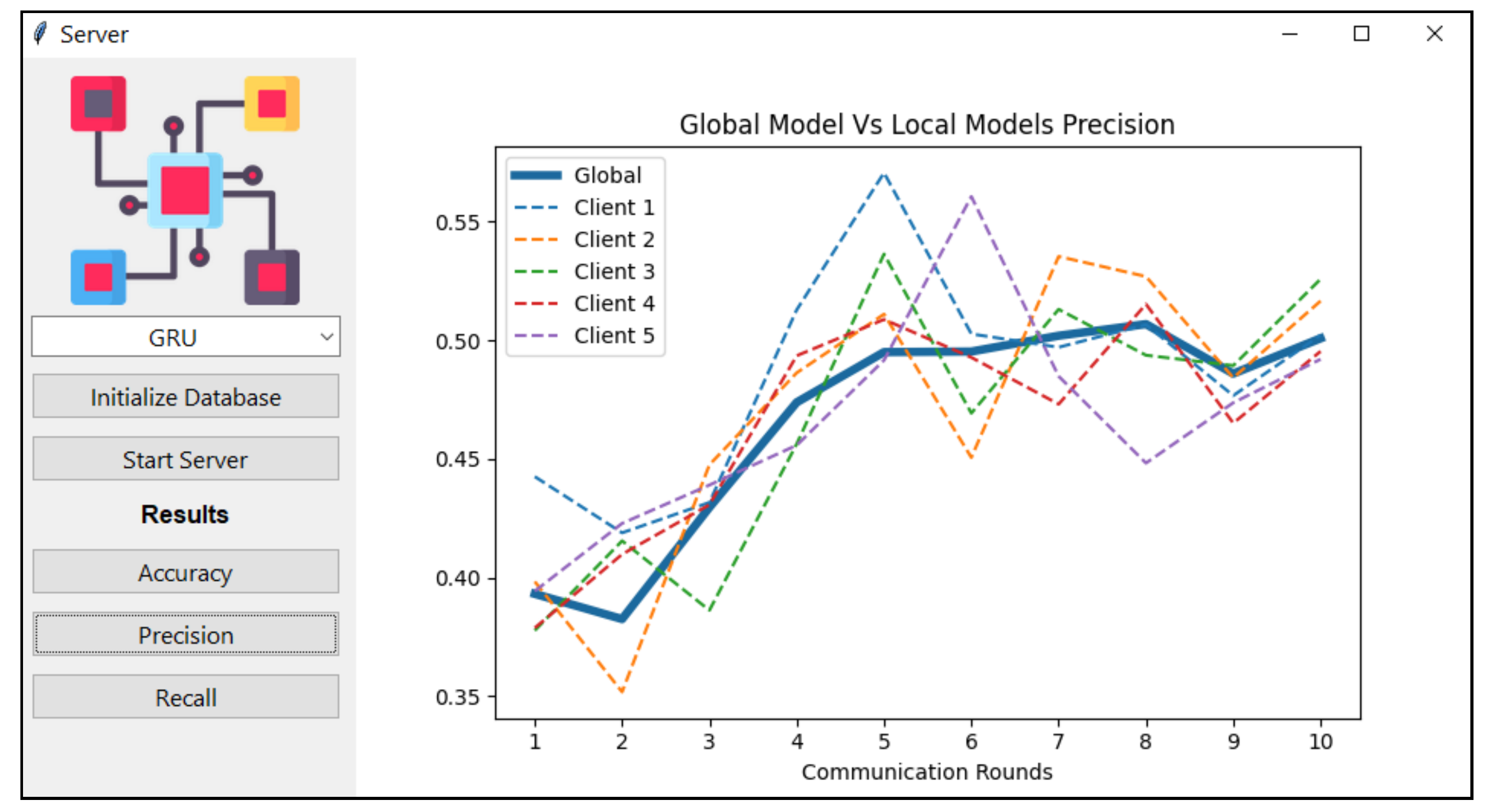}
\caption{User Interface of the Federated Learning Server App}
\label{fig5}
\end{figure}

The server application of our FL system was developed using Tkinter (8.6), and it was built with Pycharm Community Edition (2023.2.2). Unlike traditional setups where the server directly exchanges client data, our system uses FS and FRD as intermediaries. Before starting the federated learning process, the server will select the type of deep learning model to be trained, initialize a global model, and upload the model parameters to FS. The client downloads the global model from the FS to train it using its local data. This step ensures that all the clients start using the same model with the same initial parameter. After every training round, the server displays all the model's performance via the app. A screenshot of our Server app is shown in Figure \ref{fig5}.

\subsection{Evaluation Metrics}

To evaluate our model's performance, we selected Accuracy, Precision, and Recall as the evaluation metrics. Accuracy denotes the percentage of correct predictions--true positives and true negatives, out of all predictions. Precision calculates the ratio of true positives out of all positive predictions. Recall measures the ratio of true positives out of all actual positives. Although accuracy represents the overall model performance in terms of true predictions, precision and recall are crucial, particularly for imbalanced datasets, as they provide deeper insights into the results.

\begin{table*}[]
\centering
\caption{Specifications of Android Devices Used in the Experiment: Hardware and Software Details}
\label{tab:table2}
\begin{tabular}{|c|c|c|c|c|c|c|c|}
\hline
Devices & Model                                                               & \begin{tabular}[c]{@{}c@{}}Android \\ Version\end{tabular} & RAM & Chipset                                                                           & CPU                                                                                                                         & \begin{tabular}[c]{@{}c@{}}Battery \\ Capacity\\ (mAh)\end{tabular} & \begin{tabular}[c]{@{}c@{}}Battery \\ Health\end{tabular} \\ \hline
1       & \begin{tabular}[c]{@{}c@{}}Samsung Galaxy \\ S21 + 5G\end{tabular}                                                      & 14                                                         & 8GB & \begin{tabular}[c]{@{}c@{}}Qualcomm SM8350 \\Snapdragon 888 5G (5 nm)\end{tabular} & \begin{tabular}[c]{@{}c@{}}Octa-core (1x2.84 GHz, Cortex X1 \& 3x2.42 GHz, \\ Cortex A78 \& 4x1.80 GHz, Cortex A55)\end{tabular}                 & 4660                                                                & 85\%                                                      \\ \hline
2       & \begin{tabular}[c]{@{}c@{}}Samsung Galaxy \\ A24\end{tabular}       & 14                                                         & 8GB & \begin{tabular}[c]{@{}c@{}}Mediatek Helio \\ G99 (6 nm)\end{tabular}              & \begin{tabular}[c]{@{}c@{}}Octa-core (2x2.2 GHz, Cortex A76 \\ \& 6x2.0 GHz, Cortex A55)\end{tabular} & 4860                                        & 88\%                              \\ \hline
3       & \begin{tabular}[c]{@{}c@{}}Samsung Galaxy \\ A15 + 5G\end{tabular}    & 14                                                         & 8GB & \begin{tabular}[c]{@{}c@{}}MediaTek Dimensity \\ 6100 Plus\end{tabular}           & \begin{tabular}[c]{@{}c@{}}Octa-core (2.2 GHz, Dual core, Cortex A76 + \\ 2 GHz, Hexa core, Cortex A55)\end{tabular}        & 5000                                        & 90\%                              \\ \hline
4       & \begin{tabular}[c]{@{}c@{}}Samsung Galaxy \\ Tab A7\end{tabular}    & 12                                                         & 3GB & \begin{tabular}[c]{@{}c@{}}Qualcomm \\ Snapdragon 662\end{tabular}                & \begin{tabular}[c]{@{}c@{}}Octa-core (2 GHz, Quad core, Kryo 260 + \\ 1.8 GHz, Quad core, Kryo 260)\end{tabular}                               & 7040                                        & 75\%                              \\ \hline
5       & \begin{tabular}[c]{@{}c@{}}Samsung Galaxy \\ J7 Prime2\end{tabular} & 9                                                          & 3GB & \begin{tabular}[c]{@{}c@{}}Samsung Exynos 7 \\ Octa 7870\end{tabular}             & Octa-core (1.6 GHz, Cortex A53)                                                                                              & 3300                                                                & 77\%                                                      \\ \hline
\end{tabular}
\end{table*}

\begin{table}[]
\centering
\caption{Average Communication Cost of Various Models}
\label{tab:table3}
\begin{tabular}{|c|c|c|}
\hline
Model & Received Bytes (MB) & Transmitted Bytes (MB) \\ \hline
RNN        & 5.50                & 4.10                      \\ \hline
GRU        & 11.70               & 9.35                      \\ \hline
LSTM       & 14.90               & 11.85                     \\ \hline
\end{tabular}
\end{table}

\subsection{Model Performance}

The deep learning models were initially evaluated in a centralized setting to establish a performance benchmark for comparison with the federated learning setup. For the federated learning setting, we explored four distinct scenarios. The first case represents an ideal case where the data distribution among all clients is IID, and all clients participate in each training round. We mimicked a real-world situation for the next scenario by dropping one client and allowing four clients to participate in each training round. The third and fourth scenarios repeat the first two but with non-IID data distribution across clients.


\begin{table}[]
\centering
\caption{Performance of the models in centralized and federated learning setup}
\label{tab:table6}
\begin{tabular}{|c|c|c|c|c|c|}

\hline
\begin{tabular}[c]{@{}c@{}}Data\\ Distribution\end{tabular}                    & \begin{tabular}[c]{@{}c@{}}All \\ Clients\end{tabular} & Model & Accuracy  & Precision & Recall    \\ \hline
\multirow{6}{*}{\begin{tabular}[c]{@{}c@{}}Federated\\ (IID)\end{tabular}}     
                                                                               & Yes                                                    & RNN   & 0.6024    & 0.4040    & 0.3610    \\ \cline{2-6}
                                                                               & No                                                     & RNN   & 0.5835    & 0.3944    & 0.3639    \\ \cline{2-6}
                                                                               & Yes                                                    & GRU   & 0.6588    & 0.5037    & 0.4351    \\ \cline{2-6}
                                                                               & No                                                     & GRU   & 0.6637    & 0.5173    & 0.4132    \\ \cline{2-6} 
                                                                               & Yes                                                    & LSTM  &0.6570   & 0.4323    &   0.4009    \\ \cline{2-6} 
                                                                               & No                                                     & LSTM  & 0.6403    & 0.4169    & 0.3852    \\ \hline
\multirow{6}{*}{\begin{tabular}[c]{@{}c@{}}Federated\\ (Non-IID)\end{tabular}} 
                                                                               & Yes                                                    & RNN   & 0.5961    & 0.3680    & 0.3913    \\ \cline{2-6} 
                                                                               & No                                                     & RNN   & 0.5645    & 0.3547    & 0.3696    \\ \cline{2-6} 
                                                                               & Yes                                                    & GRU   & 0.6280    & 0.4765    & 0.4325    \\ \cline{2-6} 
                                                                               & No                                                     & GRU   & 0.6690    & 0.5075    & 0.4038    \\ \cline{2-6} 
                                                                               & Yes                                                    & LSTM  &   0.6670   & 0.4144   & 0.3880    \\ \cline{2-6} 
                                                                               
                                                                               & No                                                     & LSTM  & 0.6447    & 0.3912    & 0.3733    \\ \hline
\multirow{3}{*}{Centralized}                                                   & N/A                                                    & RNN   & 0.5510    & 0.4076    & 0.4125    \\ \cline{2-6} 
                                                                               & N/A                                                    & GRU   & 0.6539    & 0.4981    & 0.4774    \\ \cline{2-6} 
                                                                               & N/A                                                    & LSTM  & 0.6462    & 0.4668    & 0.4728    \\ \hline
\end{tabular}
\end{table}

For the GRU and LSTM models, performance across the first three scenarios is similar to the centralized model, with accuracy and precision remaining stable. However, we observed a notable decrease in recall, likely due to the aggregation of slightly divergent local models that may not fully capture minority class patterns. For the last scenario, where the data distribution is non-IID, and all clients do not participate in the training, there is a significant drop in the performance of all three models. It may occur because when a client containing the majority of a particular class is excluded, the models tend to overfit the remaining two classes, leading to a decline in overall performance.

Table \ref{tab:comparison} compares our best and previous studies' results. Our models demonstrated lower performance than those in earlier works because we employed standard neural networks due to limitations in computational resources in smartphones. In contrast, previous studies have utilized computationally expensive models, such as transformer based language models.

We observed that GRU outperformed LSTM by a small margin, which contradicts expectations as LSTM's complexity generally allows it to achieve superior performance. A likely reason for this result is the relatively small size of our dataset, which is inadequate for completely training LSTMs. GRUs, being simpler models, converged faster and more effectively.

\begin{figure*}[!t]
\centering
\captionsetup[subfigure]{font=scriptsize}
\subfloat[Accurcay Vs Communication Rounds]{\includegraphics[width=0.33\linewidth]{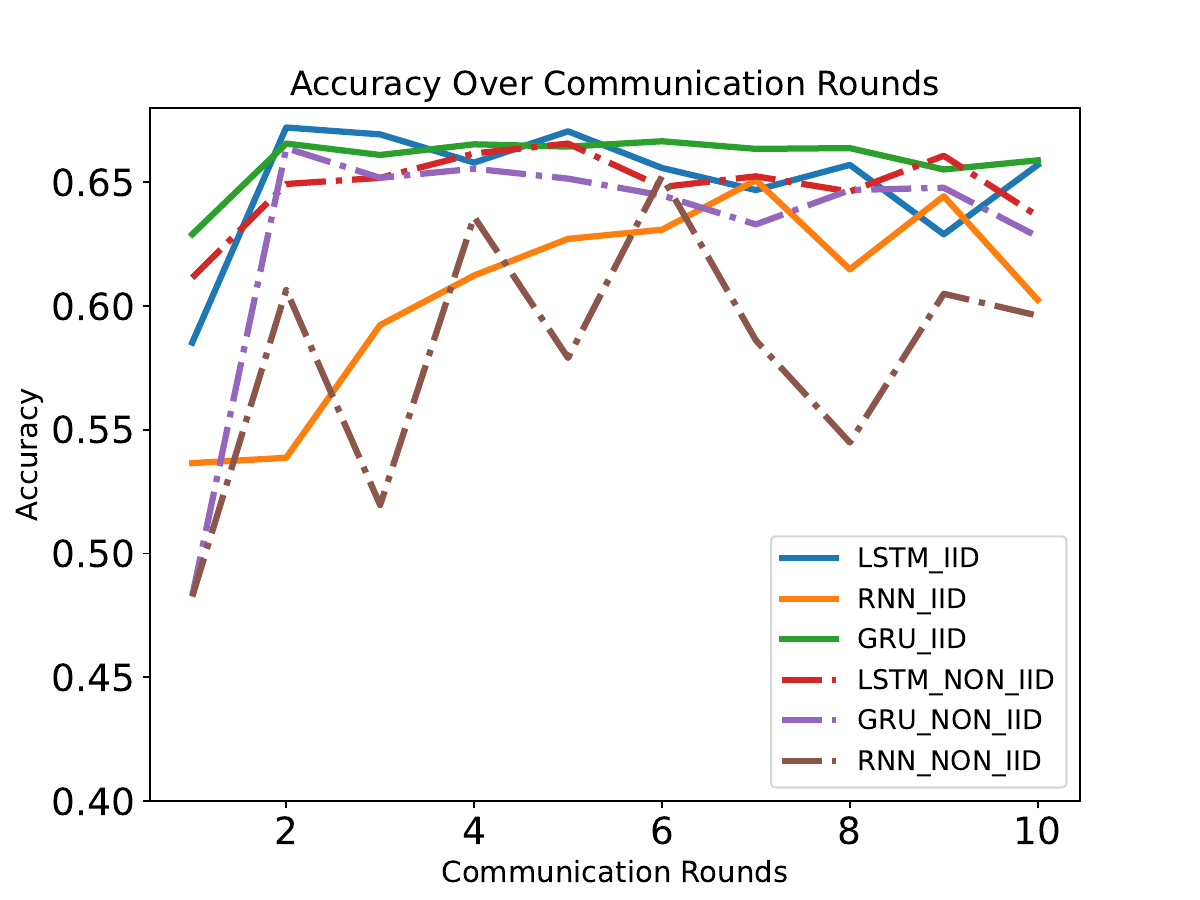}%
\label{fig6_first_case}}
\hfil
\subfloat[Precision Vs Communication Rounds]{\includegraphics[width=0.33\linewidth]{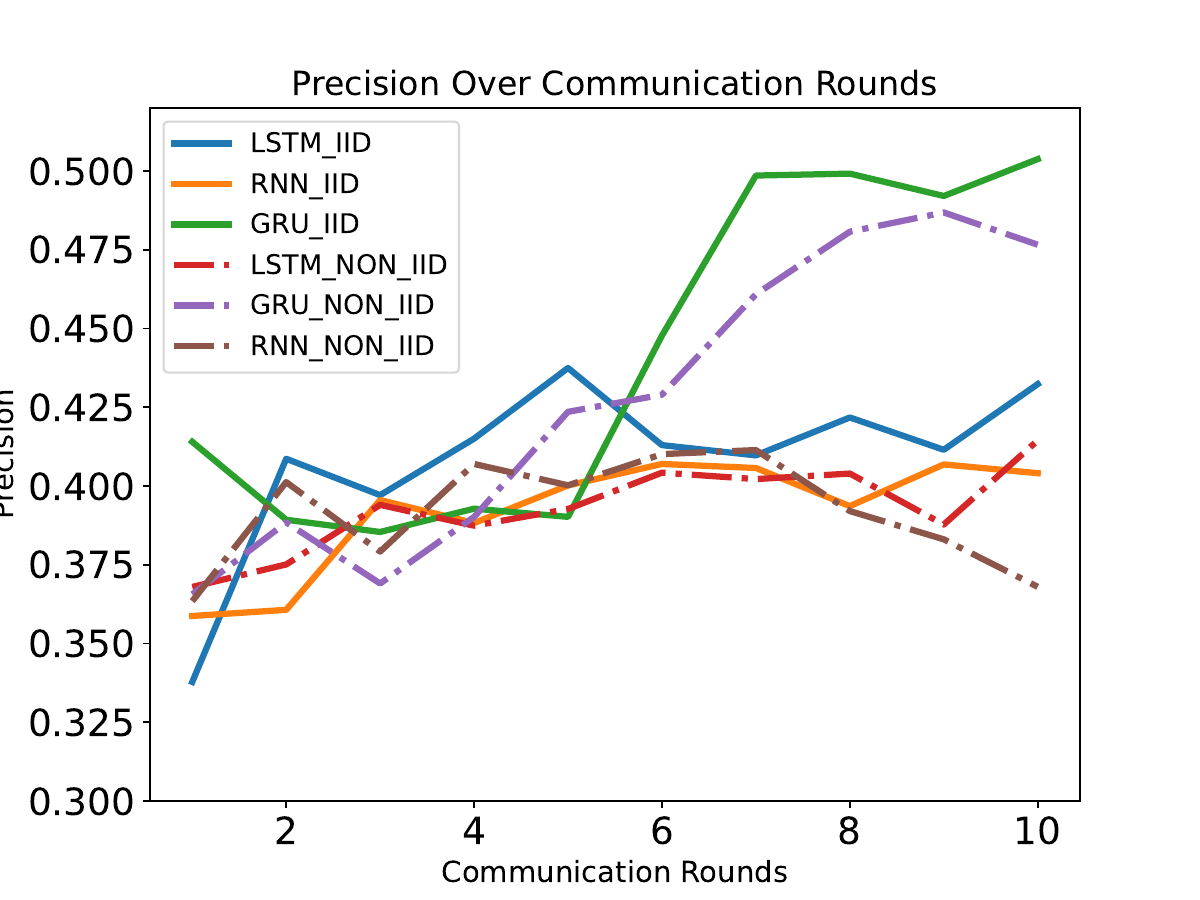}%
\label{fig6_second_case}}
\hfil
\subfloat[Recall Vs Communication Rounds]{\includegraphics[width=0.33\linewidth]{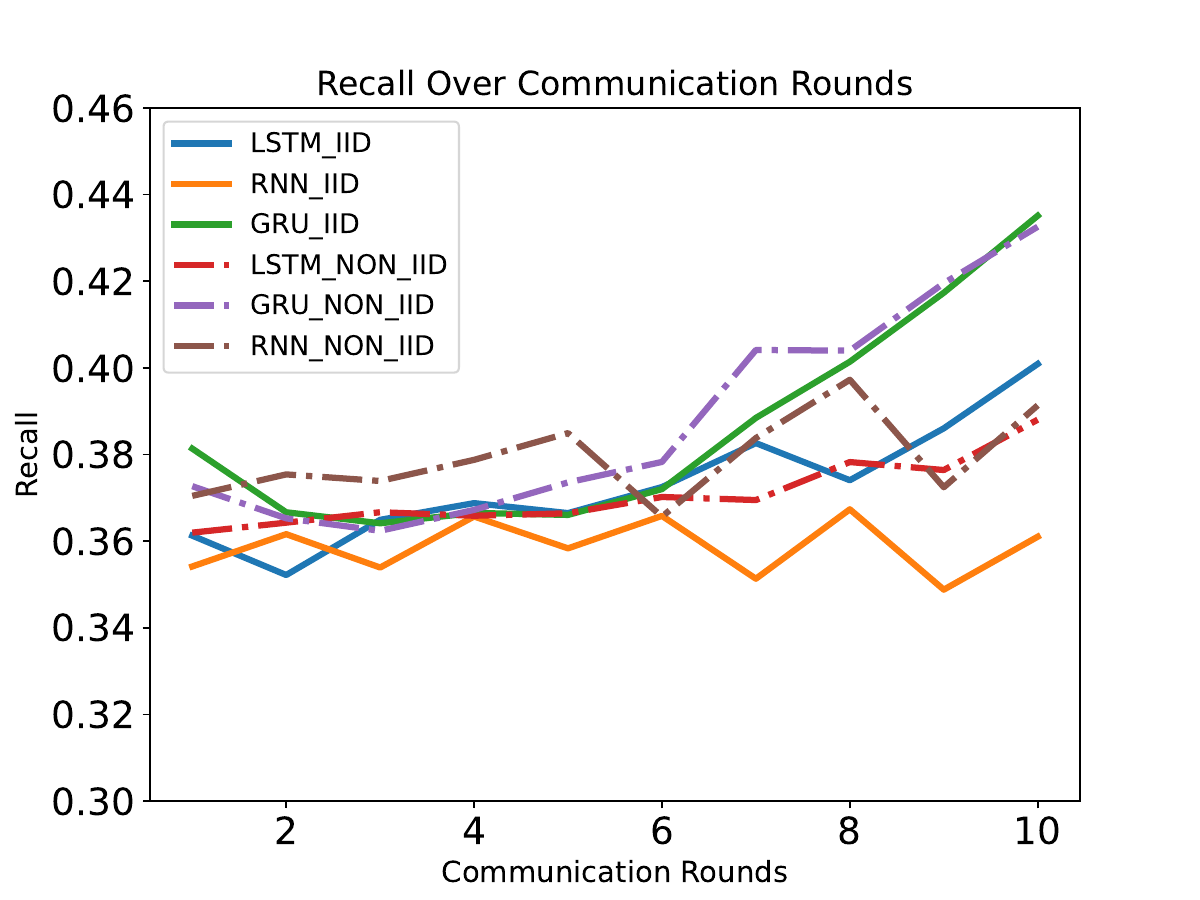}
\label{fig6_third_case}}
\caption{Performance Analysis of the Global Models}
\label{fig6}
\end{figure*}

\begin{table}[]
\centering
\caption{Comparison of Our Best Model with Previous Studies in Centralized Settings}
\label{tab:comparison}
\begin{tabular}{|c|c|c|c|}
\hline
Work Approach                                                                                                                                                                  & Accuracy & Precision & Recall \\ \hline
Our GRU model (Centralized) & 0.6539   & 0.4981    & 0.4774 \\ \hline
Our GRU model (Federated) & 0.6588 & 0.5037 &  0.4351 \\ \hline
\begin{tabular}[c]{@{}c@{}} Ensemble  transformer-based \\language models \cite{poswiata2022opi}\end{tabular} & 0.6582   & 0.5860  & 0.5912 \\ \hline
\begin{tabular}[c]{@{}c@{}}Ensemble  transformer-based\\ language models \cite{wang2022nycu_twd}\end{tabular}  & 0.6330  & 0.5394  & 0.5732 \\ \hline
Knowledge Graph \cite{tavchioski2022e8}       & 0.6015   & 0.5149    & 0.5714 \\ \hline

RoBERTa \cite{sivamanikandan2022scubemsec}        & 0.5106   & 0.4610    & 0.5193 \\ \hline
Logistic Regression \cite{agirrezabal2022kucst}     & 0.5464   & 0.4321    & 0.4728 \\ \hline
\end{tabular}
\end{table}

\subsection{Communication Cost}
The internet plays a vital role in federated learning by serving as the medium that connects devices. Therefore, analyzing network activity during federated learning training is essential. Table \ref{tab:table3} shows the average data transmitted and received per communication round. This data transfer includes model uploads and downloads from FS and exchanged messages with the FRD. When we upload or download models from FS, it implies that the weights and biases of each layer of the deep learning models are being transferred. Additionally, we observe that higher average data is transferred as the model's complexity increases due to the larger number of parameters in more complex models.

Interestingly, although the file sizes for local and global models are identical,  Table \ref{tab:table3}  reveals that the received bytes are consistently higher than the transmitted bytes. This discrepancy originates from devices that finish early. When a device finishes early, it periodically checks whether the server has uploaded a new global model. This repeated polling inflates the average amount of received data.

\subsection{Client Device Time Profiling}

Table \ref{tab:table2} outlines the hardware and software specifications of the devices used in our study. The device selection is a range of models, from older devices, such as Device 5, to newer ones, such as Device 1. Device 3, a tablet, performs comparably to the other devices. Although we have only shown results from Samsung devices for fair comparison, we also conducted experiments using Xiaomi and OnePlus smartphones and observed similar results.

\subsubsection{Training Time}

Figure \ref{fig7} shows the training time of the three deep learning models. Our findings are consistent with the models' theoretical complexities. LSTM requires the longest time to train due to its complex gating mechanism. Following LSTM, GRU is the next in line, has a more streamlined gating mechanism, and takes less time to train than LSTMs. RNNs take the least training time as they have a straightforward recurrence mechanism that lacks a gating structure like GRUs and LSTMs.

Device 1 (Samsung Galaxy S21 + 5G) was the fastest due to its powerful Qualcomm Snapdragon 888 chipset, which has a high-performance Cortex-X1 core clocked at 2.84 GHz and 8GB of RAM, allowing it to handle deep learning tasks efficiently. Device 5 (Samsung Galaxy J7 Prime2) was the slowest, especially because of its older Exynos 7870 chipset, using low-power Cortex-A53 cores running at 1.6 GHz, and just 3GB of RAM, which limited its capability to process deep learning models effectively. Devices 2 (Samsung Galaxy A24) and 3 (Samsung Galaxy A15 + 5G), despite having 8GB of RAM, were constrained by their mid-range MediaTek chipsets, with only two Cortex-A76 performance cores and six efficiency Cortex-A55 cores, which reduced their ability to match the flagship-level performance of the Snapdragon 888. Device 4 (Samsung Galaxy Tab A7) performed better than Device 5 but was still slower due to its Snapdragon 662 chipset, lower clock speeds (2 GHz), and limited 3GB of RAM. Overall, training time differences were driven by the power of the chipsets, CPU architecture, and RAM.

\subsubsection{Overhead Time}
In addition to training the models, client devices must perform several other tasks, which include loading the training data, tokenizer, and embedding files. The raw training data is processed and prepared for input into the neural network. Collectively, these activities add to the overhead time. It took nearly 0.5 to 2 seconds to perform these procedures for nearly all devices, except for device 5, which required around 3 to 4 seconds.

\subsubsection{Upload \& Download Time}
\begin{figure*}[!t]
\centering
\captionsetup[subfigure]{font=scriptsize}
\subfloat[RNN]{\includegraphics[width=0.3333\linewidth]{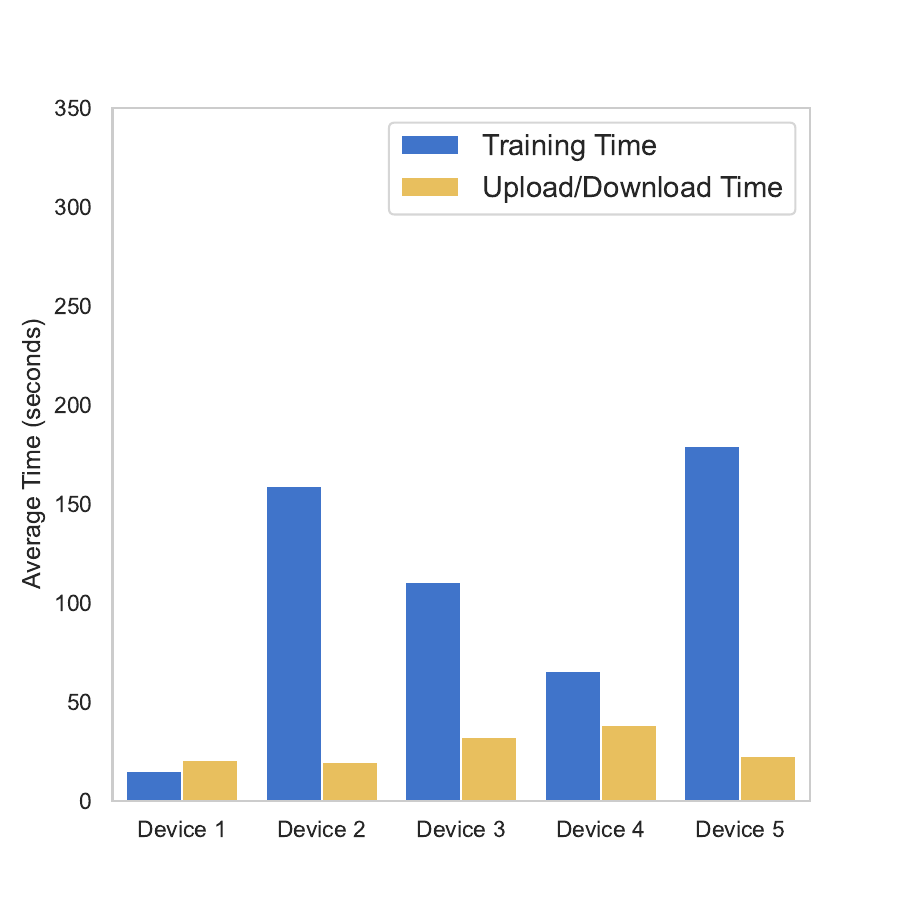}%
\label{fig7_second_case}}
\hfil
\subfloat[GRU]{\includegraphics[width=0.3333\linewidth]{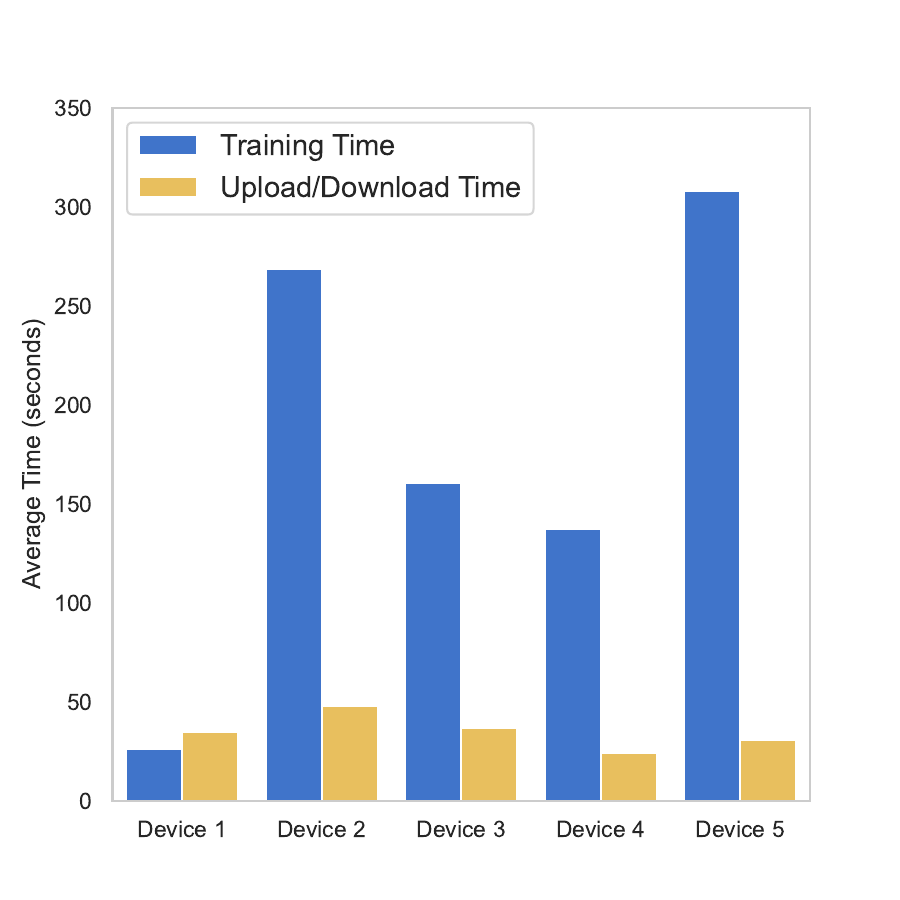}%
\label{fig7_first_case}}
\hfil
\subfloat[LSTM]{\includegraphics[width=0.3333\linewidth]{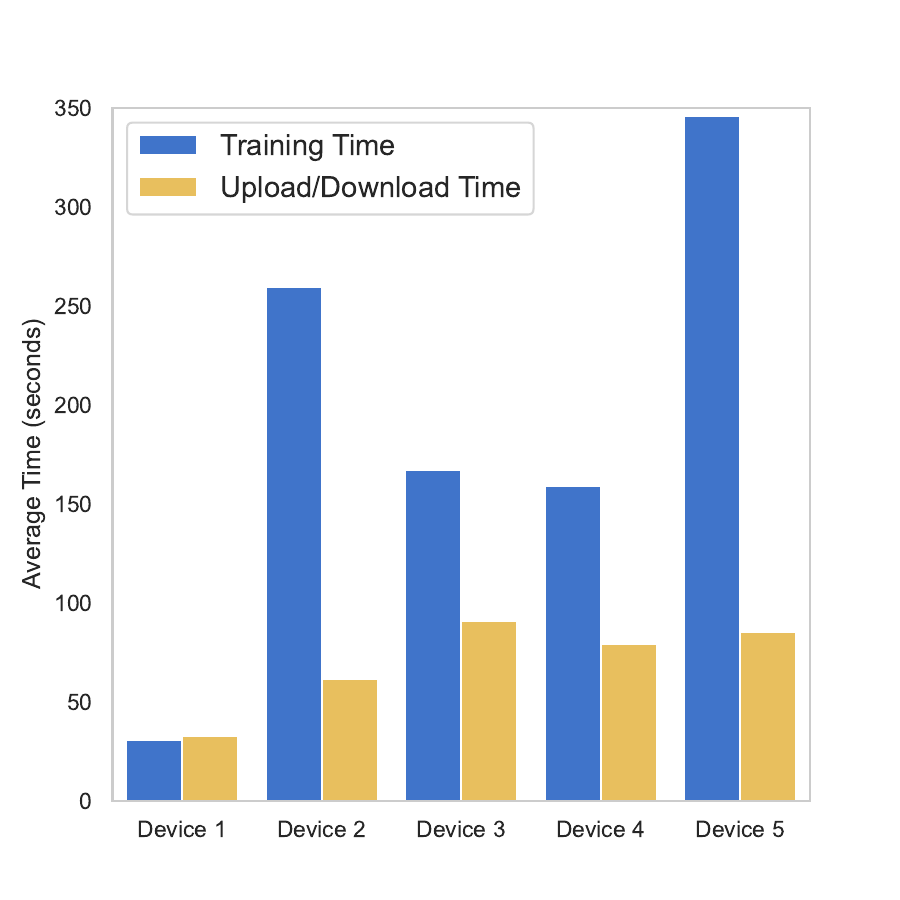}%
\label{fig7_third_case}}
\caption{Training Time and Upload \& Download Time of the Neural Network Models}
\label{fig7}
\end{figure*}

For each training round, the client devices must send the locally trained parameters and retrieve global parameters to and from FS, contributing to the upload and download time. Figure \ref{fig7} reveals that, unlike training time, there is no discernible trend in upload and download time. The devices were geographically distributed and connected to different networks. Devices 1, 3, and 4 connected via Wi-Fi, while Devices 2 and 5 connected to mobile networks. Different networks have varying speeds, leading to variability in data transfer times.

\subsubsection{Inference Time}

Inference time refers to the time a model takes to make predictions for a given sample, and it was determined by averaging the time required to predict 100 samples from the training dataset. The inference time of the devices is shown in Table \ref{tab:table8}. We can observe a trend similar to training time, where faster devices result in shorter inference time.

\subsection{Battery Usage}

We tracked the battery level before and after each training round to assess the average battery consumption per round. The total battery capacity of Android devices diminishes depending on usage. We used a third-party app, AccuBattery \cite{googleAccuBatteryApps}, to estimate each device's effective battery capacity. The battery health of each device is shown in Table \ref{tab:table2}. The total Milliampere-hour (mAh) consumed per round was calculated by multiplying the average battery level drop by the effective battery capacity. Table \ref{tab:table10} indicates this as the model complexity rises correspondingly. Furthermore, we can also infer that lower battery health results in a more significant battery drop.


\begin{table}[]
\centering
\caption{Inference time (milliseconds) for a sample}
\label{tab:table8}
\begin{tabular}{|c|ccccc|}
\hline
\multirow{2}{*}{Model} & \multicolumn{5}{c|}{Devices Number}                                                                                                      \\ \cline{2-6} 
                       & \multicolumn{1}{c|}{Device 1} & \multicolumn{1}{c|}{Device 2} & \multicolumn{1}{c|}{Device 3} & \multicolumn{1}{c|}{Device 4} & Device 5 \\ \hline
RNN                    & \multicolumn{1}{c|}{3}        & \multicolumn{1}{c|}{19}       & \multicolumn{1}{c|}{15}       & \multicolumn{1}{c|}{12}       & 24       \\ \hline
GRU                    & \multicolumn{1}{c|}{6}        & \multicolumn{1}{c|}{45}       & \multicolumn{1}{c|}{30}       & \multicolumn{1}{c|}{29}       & 65       \\ \hline
LSTM                   & \multicolumn{1}{c|}{7}        & \multicolumn{1}{c|}{59}       & \multicolumn{1}{c|}{40}       & \multicolumn{1}{c|}{31}       & 73       \\ \hline
\end{tabular}
\end{table}


\begin{table}[]
\centering
\caption{Average battery usage (mAh) per round}
\label{tab:table10}
\begin{tabular}{|c|ccccc|}
\hline
\multirow{2}{*}{Model} & \multicolumn{5}{c|}{Devices Number}                                                                                                      \\ \cline{2-6} 
                       & \multicolumn{1}{c|}{Device 1} & \multicolumn{1}{c|}{Device 2} & \multicolumn{1}{c|}{Device 3} & \multicolumn{1}{c|}{Device 4} & Device 5 \\ \hline
RNN                    & \multicolumn{1}{c|}{15}       & \multicolumn{1}{c|}{12}       & \multicolumn{1}{c|}{11}       & \multicolumn{1}{c|}{52}       & 18       \\ \hline
GRU                    & \multicolumn{1}{c|}{26}       & \multicolumn{1}{c|}{22}       & \multicolumn{1}{c|}{22}       & \multicolumn{1}{c|}{70}       & 38       \\ \hline
LSTM                   & \multicolumn{1}{c|}{36}       & \multicolumn{1}{c|}{35}       & \multicolumn{1}{c|}{36}       & \multicolumn{1}{c|}{84}       & 60       \\ \hline
\end{tabular}
\end{table}

\section{Limitations}
A key limitation of our study is that our federated learning operates synchronously. If any device finished early, it had to wait for the slowest device to complete its training, which was a bottleneck. This issue can be resolved by incorporating asynchronous approaches \cite{jiang2022flsys}\cite{katevas2022flaas}. Another problem we faced with Firebase was that on rare occasions, when the clients tried to load the global model weights, Firebase would not respond, causing the app to stall. In this case, we had to restart that particular client device manually. Furthermore, we included Python libraries, such as Tensorflow, Matplotlib, Pandas, and Numpy in our app, which resulted in a significantly larger Android app size.  

\section{Conclusion}
We present a Federated Learning based approach for detecting depression levels from Reddit posts. Unlike previous studies that mainly focus on simulation-based FL scenarios, we utilize smartphones as local devices and demonstrate results in a real-world environment. To optimize text processing, we employed a common tokenizer across all clients. We assess the performance of multiple deep learning architectures, including GRU, RNN, and LSTM.  Additionally, we consider a significant challenge in FL--non-IID data distribution. We achieve a promising accuracy of 66\% with the GRU architecture in the federated setting, highlighting the potential of our approach for depression detection. Moreover, we provide a comprehensive analysis of resource consumption by monitoring network and battery usage, as well as performing client device time profiling. Our future work will focus on refining the current method to handle non-IID distribution better and incorporating techniques such as personalization to support individualized mental health prediction. Additionally, we aim to incorporate differential privacy to further enhance data protection and ensure stronger privacy guarantees.

\bibliographystyle{IEEEtran}
\bibliography{bare_jrnl_new_sample4}

\vfill

\end{document}